# Color Image Enhancement Using the *lrgb* Coordinates in the Context of Support Fuzzification

**Vasile Pătraşcu**

Department of Informatics Technology
TAROM Company
Şos. Bucureşti-Ploieşti, K16.5, Bucureşti-Otopeni, Romania
e-mail: vpatrascu@tarom.ro

**Abstract:** Image enhancement is an important stage in the image-processing domain. The most known image enhancement method is the histogram equalization. This method is an automated one, and realizes a simultaneous modification for brightness and contrast in the case of monochrome images and for brightness, contrast, saturation and hue in the case of color images. Simple and efficient methods can be obtained if affine transforms within logarithmic models are used. A very important thing in the affine transform determination for color images is the coordinate system that is used for color space representation. Thus, the using of the *RGB* coordinates leads to a simultaneous modification of luminosity and saturation. In this paper using the *lrgb* perceptual coordinates one can define affine transforms, which allow a separated modification of luminosity *l* and saturation *s* (saturation being calculated with the component *rgb* in the chromatic plane). Better results can be obtained if partitions are defined on the image support and then the pixels are separately processed in each window belonging to the defined partition. Classical partitions frequently lead to the appearance of some discontinuities at the boundaries between these windows. In order to avoid all these drawbacks the classical partitions may be replaced by fuzzy partitions. Their elements will be fuzzy windows and in each of them there will be defined an affine transform induced by parameters using the fuzzy mean, fuzzy variance and fuzzy saturation computed for the pixels that belong to the analyzed window. The final image is obtained by summing up in a weight way the images of every fuzzy window.

## 1  Introduction

In this paper a new enhancement method for color images is presented. One of the most known enhancement methods is the histogram equalization. The histogram equalization method belongs to the non-linear transform category. In the case of monochrome images, the resulted image is very close (from statistical point of view) to an image having a uniform distribution of its intensity values. Similar results can be obtained using the logarithmic affine transforms [1,2,3,6]. One can determine the affine transform parameters taking into account the image statistics properties and using the mean and variance. In the necessary calculus, the utilization of real number algebra operations may lead to some results that are outside of image values domain. A way to avoid this drawback is the utilization of algebrical structures defined on real and bounded sets. The determination of the affine transform parameters is made taking into account the statistics of luminosity. The affine transform with two parameters can be seen as a sequence of two steps: a translation and then a scalar multiplication. The translation leads to a brightness modification for both monochrome and color images. The scalar multiplication leads to a contrast modification for monochrome images and a simultaneous modification of contrast luminosity and saturation for color images. In order to separate these two modifications there will be used an affine transform with three parameters having the following functions: the first determines the translation and is computed using the luminosity average; the second is a multiplication factor for luminosity and is determined using the luminosity variance; the third is also a multiplication factor only for chromatic components and is determined by the average of image saturation. In order to compute these three parameters it was necessary to use a perceptual system (called *lrgb*) instead of the *RGB* one [8]. Better results can be obtained if partitions are defined on the image support and then the pixels are separately processed in each window belonging to the defined partition [4,5]. The classical partitions frequently lead to the appearance of some discontinuities at the boundaries between these windows. In order to avoid all these drawbacks the classical partitions may be replaced by fuzzy partitions. Their elements will be fuzzy windows and in each of them there will be defined an affine transform induced by parameters using the fuzzy mean, fuzzy variance and fuzzy saturation computed for the pixels that belong (in fuzzy meaning) to the analyzed window. Their calculus uses logarithmical operations. The final image is obtained by summing up in a weight way the images of every fuzzy window as defined before. Within this merging operation, the weights used are membership degrees, which define the fuzzy partition of the image support. The paper has the following structure: section 2 comprises a short presentation of the logarithmic model used for the calculus; section 3 presents the perceptual coordinate system called *lrgb* system; section 4 presents the affine transform for monochrome and color images; section 5 presents the fuzzy partition on the image support and then, the determination of the affine transforms for each component (window) of the partition is presented in section 6. Some experimental results are shown in section 7. Finally, section 8 comprises a few conclusions.

## 2  The Logarithmic Model

Let us consider the set $V = (0,1)$ as the space of intensity values [7]. The addition $\langle + \rangle$ and the multiplication $\langle \times \rangle$ by a real scalar will be defined in the set *V*. Then, defining a scalar product $(\cdot | \cdot)_V$ and a norm $\|.\|_V$, a Euclidean space will be defined.

We will consider the addition:

$\forall v_1, v_2 \in V$,

$$v_1 \langle + \rangle v_2 = \frac{v_1 v_2}{(1-v_1)(1-v_2) + v_1 v_2}$$

The neutral element for the addition is $\theta = 0.5$. Each element $v \in V$ has an opposite $w = 1 - v$.

The subtraction operation $\langle - \rangle$ is defined by:

$\forall v_1, v_2 \in V$,

$$v_1 \langle - \rangle v_2 = \frac{v_1(1-v_2)}{v_1(1-v_2) + (1-v_1)v_2}$$

The scalar multiplication is defined by:

$\forall v \in V, \forall \lambda \in R$,

$$\lambda\langle\times\rangle v = \frac{v^{\lambda}}{v^{\lambda}+(1-v)^{\lambda}}$$

The vector space of intensity values $(V,\langle+\rangle,\langle\times\rangle)$ is isomorphic to the space of real numbers $(R,+,\cdot)$ by the function $\varphi:V\rightarrow R$, defined as:

$\forall v\in V$,

$$\varphi(v)=\frac{1}{4}\ln\left(\frac{v}{1-v}\right) \qquad (2.1)$$

The isomorphism $\varphi$ verifies:

$\forall v_1, v_2 \in V$,

$$\varphi(v_1\langle+\rangle v_2)=\varphi(v_1)+\varphi(v_2)$$

$\forall\lambda\in R, \forall v\in V$,

$$\varphi(\lambda\langle\times\rangle v)=\lambda\cdot\varphi(v)$$

The scalar product $(\cdot\,|\,\cdot)_V:V\times V\rightarrow R$ is defined using the isomorphism (2.1) as:

$\forall v_1, v_2 \in V$,

$$(v_1\,|\,v_2)_V=\varphi(v_1)\cdot\varphi(v_2)$$

Based on the scalar product $(\cdot\,|\,\cdot)_V$ the vector space $V$ becomes a Euclidean space.

The norm $\|\cdot\|_V:V\rightarrow R^+$ is defined via the scalar product:

$\forall v\in V$,

$$\|v\|_V=\sqrt{(v\,|\,v)_V}=|\varphi(v)|$$

This norm verifies the following relations:

$$\|v_1\langle+\rangle v_2\|_V=|\varphi(v_1)+\varphi(v_2)|$$

$$\|v_1\langle-\rangle v_2\|_V=|\varphi(v_1)-\varphi(v_2)|$$

The modulus $|\,.\,|_V:V\rightarrow V^+=[\theta,1]$ defined by:

$\forall v\in V$,

$$|v|_V=0.5+|v-0.5|$$

verifies the triangle inequality:

$\forall v_1, v_2 \in V$,

$$|v_1|_V\,\langle+\rangle\,|v_2|_V\geq|v_1\langle+\rangle v_2|_V$$

## 3  The Coordinate System *lrgb*

### 3.1  The Coordinate System *lrgb* Definition Using Classical Operations

One can define the perceptual coordinate system $l, r, g, b$ (where *l, r, g, b* are measures of *whiteness, redness, greenness* and *blueness* of the image) starting from *RGB* system and using the following formulae:

$$l = \frac{R + G + B}{3} \qquad (3.1.1)$$

$$r = \frac{2R - G - B}{3} \qquad (3.1.2)$$

$$g = \frac{2G - B - R}{3} \qquad (3.1.3)$$

$$b = \frac{2B - R - G}{3} \qquad (3.1.4)$$

One defines the color saturation by:

$$s^2(r, g, b) = \frac{r^2 + g^2 + b^2}{3}$$

The following relations make the passing from *lrgb* system to the *RGB* one:

$$R = l + \frac{2r - g - b}{3}$$

$$G = l + \frac{2g - b - r}{3}$$

$$B = l + \frac{2b - r - g}{3}$$

### 3.2  The Coordinate System *lrgb* Definition Using Logarithmic Operations

Starting from relations (3.1.1-3.1.4) one can obtain new relations using logarithmic operations instead of classical ones:

$$l = \frac{1}{3} \langle \times \rangle \left( R \langle + \rangle G \langle + \rangle B \right) \qquad (3.2.1)$$

$$r = \frac{1}{3} \langle \times \rangle \left( 2 \langle \times \rangle R \langle - \rangle G \langle - \rangle B \right) \qquad (3.2.2)$$

$$g = \frac{1}{3} \langle \times \rangle \left( 2 \langle \times \rangle G \langle - \rangle B \langle - \rangle R \right) \qquad (3.2.3)$$

$$b = \frac{1}{3} \langle \times \rangle \left( 2 \langle \times \rangle B \langle - \rangle R \langle - \rangle G \right) \qquad (3.2.4)$$

One defines the color saturation by:

$$s_\varphi^2(r, g, b) = \frac{\varphi^2(r) + \varphi^2(g) + \varphi^2(b)}{3}$$

# 4 The Affine Transforms on the Image Set

## 4.1 The Monochrome and Color Images

A monochrome or scalar image is described by a real and bounded function $f : \Omega \to V$, where $\Omega \subset R^2$ is a compact set that represents the image support and $V \subset R$ is the bounded set of values used for image representation. The interval $V = (0,1)$ was considered as values set in this paper. A color image is described in the *RGB* coordinate system by three scalar functions: $f_R : \Omega \to V$, $f_G : \Omega \to V$, $f_B : \Omega \to V$ that define the *red, green, blue* components of the color. The monochrome image set will be noted with $F(\Omega, V)$ and the color image set with $F(\Omega, V^3)$.

## 4.2 The Affine Transform on the Monochrome Images

Let us consider these affine transforms on the monochrome images $F(\Omega, V)$ defined as following: $\psi : F(\Omega, V) \to F(\Omega, V)$,

$$\psi(f) = \lambda \langle \times \rangle (f \langle + \rangle \tau) \qquad (4.2.1)$$

where $\lambda \in R$ and $\tau \in V$. This form shows that an image can be processed in two steps: a translation with a constant value $\tau$, which leads to a change in the image brightness, then a scalar multiplication with the factor $\lambda$ - leading to a change in the image contrast. The parameters are chosen in such a way to get a new image very close (from a statistical point of view) to an image *u* with a uniform distribution of its intensity values on the set *V*. This criterion shows that the enhanced image must have its mean $\mu(u) = 0.5$ and its variance $\sigma^2(u) = 1/12$, which are exactly the values owned by a random variable with a uniform distribution on the interval $[0,1]$. In fact, as a result, we approximate the nonlinear transform yielded by the algorithm of histogram equalization with an affine transform as the one in (4.2.1). In these conditions for any image $f$ with the mean $\mu_\varphi(f)$ and the variance $\sigma^2_\varphi(f)$, the affine transform $\psi$ becomes:

$$\psi(f) = \frac{\sigma(u)}{\sigma_\varphi(f)} \langle \times \rangle \left( f \langle - \rangle \mu_\varphi(f) \right)$$

The mean and variance are defined by:

$$\mu_\varphi(f) = \frac{1}{card(\Omega)} \langle \times \rangle \left( \underset{\Omega}{\langle + \rangle} f \right)$$

$$\sigma^2_\varphi(f) = \frac{1}{card(\Omega)} \sum_\Omega \left\| f \langle - \rangle \mu_\varphi(f) \right\|^2_V$$

### 4.3  The Affine Transforms on Color Images Using Two Parameters

The two parameter affine transform for color images are defined by the following relations:

$$R_{enh} = \lambda \langle \times \rangle (R \langle + \rangle \tau) \tag{4.3.1}$$

$$G_{enh} = \lambda \langle \times \rangle (G \langle + \rangle \tau) \tag{4.3.2}$$

$$B_{enh} = \lambda \langle \times \rangle (B \langle + \rangle \tau) \tag{4.3.3}$$

where

$$\lambda = \frac{\sigma(u)}{\sigma_\varphi(l)}$$

$$\tau = \langle - \rangle \mu_\varphi(l)$$

and $l$ is the luminosity (3.2.1).

One can see that the same affine transform has been used for each color component

### 4.4  The Affine Transforms on Color Images Using Three Parameters

Using the coordinates *lrgb* the relations (4.3.1, 4.3.2, 4.3.3) become:

$$R_{enh} = \lambda \langle \times \rangle (l \langle + \rangle \tau) \langle + \rangle \lambda \langle \times \rangle r \tag{4.4.1}$$

$$G_{enh} = \lambda \langle \times \rangle (l \langle + \rangle \tau) \langle + \rangle \lambda \langle \times \rangle g \tag{4.4.2}$$

$$B_{enh} = \lambda \langle \times \rangle (l \langle + \rangle \tau) \langle + \rangle \lambda \langle \times \rangle b \tag{4.4.3}$$

From (4.4.1, 4.4.2, 4.4.3) it results that:

$$l_{enh} = \lambda \langle \times \rangle (l \langle + \rangle \tau)$$

$$s_\varphi(r_{enh}, g_{enh}, b_{enh}) = \lambda \cdot s_\varphi(r, g, b) \tag{4.4.4}$$

In relation (4.4.4) one can see that the saturation of the original image is multiplied by parameter $\lambda$ that depends on luminosity variance. For this reason it is necessary to introduce the third parameter $\omega$ that will be calculated using the chromatic components. Thus the affine transforms will have the following forms:

$$R_{enh} = \lambda \langle \times \rangle (l \langle + \rangle \tau) \langle + \rangle \omega \langle \times \rangle r$$

$$G_{enh} = \lambda \langle \times \rangle (l \langle + \rangle \tau) \langle + \rangle \omega \langle \times \rangle g$$

$$B_{enh} = \lambda \langle \times \rangle (l \langle + \rangle \tau) \langle + \rangle \omega \langle \times \rangle b$$

where

$$\omega = \frac{\sigma(u)}{\gamma_\varphi(r, g, b)}$$

and

$$\gamma_\varphi^2(r, g, b) = \frac{1}{card(\Omega)} \cdot \sum_\Omega s_\varphi^2(r, g, b)$$

is the image saturation.

## 5  The Fuzzification of the Image Support

Let be $\Omega \subset R^2$ an image support. Without losing generality of the problem, the rectangle $\Omega = [x_0, x_1] \times [y_0, y_1]$ can be considered as an image support. On the set $\Omega$ a fuzzy partition is built and its elements are called fuzzy windows. The coordinates of a pixel within the support $\Omega$ will be noted $(x, y)$. Let there be $P = \left\{ W_{ij} \mid (i, j) \in [0, m] \times [0, n] \right\}$ a fuzzy partition of the support $\Omega$. Consider for $(i, j) \in [0, m] \times [0, n]$ the polynomials $p_{ij} : \Omega \to [0,1]$,

$$p_{ij}(x, y) = C_m^i C_n^j \frac{(x - x_0)^i (x_1 - x)^{m-i}}{(x_1 - x_0)^m} \cdot \frac{(y - y_0)^j (y_1 - y)^{n-j}}{(y_1 - y_0)^n}$$

where $C_m^i = \dfrac{m!}{i!(m-i)!}$, $C_n^j = \dfrac{n!}{j!(n-j)!}$. The membership degrees of a point $(x, y) \in \Omega$ to the fuzzy window $W_{ij}$ are given by the functions $w_{ij} : \Omega \to [0,1]$ defined by the relation:

$$w_{ij}(x, y) = \frac{\left( p_{ij}(x, y) \right)^{\gamma}}{\sum\limits_{j=0}^{n} \sum\limits_{i=0}^{m} \left( p_{ij}(x, y) \right)^{\gamma}} \tag{5.1}$$

The parameter $\gamma \in (0, \infty)$ has the role of a tuning parameter offering a greater flexibility in building the fuzzy partition $P$. In other words $\gamma$ controls the fuzzification-defuzzification degree of the partition. The membership degrees $w_{ij}(x, y)$ describe the position of the point $(x, y)$ within the support $\Omega$, namely the upper part or the lower part, the left hand part, the right hand part or in the center of the image. For each window $W_{ij}$ defined by (5.1), the fuzzy cardinality is computed as follows:

$$card(W_{ij}) = \sum_{\Omega} w_{ij}$$

Further on, fuzzy statistics of an image are used in relation to the window $W_{ij}$. For a monochrome image $f$ the fuzzy mean $\mu_{\varphi}(f, W_{ij})$ and the fuzzy variance $\sigma_{\varphi}^2(f, W_{ij})$ of the image $f$ within window $W_{ij}$ are defined by:

$$\mu_{\varphi}(f, W_{ij}) = \left\langle + \right\rangle_{\Omega} \left( \frac{w_{ij}}{card(W_{ij})} \langle \times \rangle f \right)$$

$$\sigma_\varphi^2(f,W_{ij}) = \sum_\Omega \frac{w_{ij}\parallel f\langle-\rangle\mu_\varphi(f,W_{ij})\parallel_V^2}{card(W_{ij})}$$

For a color image having its four-*lrgb* components, fuzzy mean $\mu_\varphi(l,W_{ij})$, fuzzy variance $\sigma_\varphi^2(l,W_{ij})$ and fuzzy saturation $\gamma_\varphi^2(r,g,b,W_{ij})$ within window $W_{ij}$ are defined by:

$$\mu_\varphi(l,W_{ij}) = \left\langle + \right\rangle_\Omega \left( \frac{w_{ij}}{card(W_{ij})}\langle\times\rangle l \right) \qquad (5.2)$$

$$\sigma_\varphi^2(l,W_{ij}) = \sum_\Omega \frac{w_{ij}\parallel l\langle-\rangle\mu_\varphi(l,W_{ij})\parallel_V^2}{card(W_{ij})} \qquad (5.3)$$

$$\gamma_\varphi^2(r,g,b,W_{ij}) = \frac{1}{card(W_{ij})}\cdot\sum_\Omega w_{ij}\cdot s_\varphi^2(r,g,b) \qquad (5.4)$$

## 6  The Image Enhancement Methods

### 6.1  The Enhancement Method for Monochrome  Images

In the context of support fuzzification an affine transform will be computed for each element. The fuzzy window $W_{ij}$ will supply a couple of parameters $(\lambda,\tau)$, which reflects the statistics according to the pixels belonging (in fuzzy meaning) to this window. Thus

$$\lambda_{ij} = \frac{\sigma(u)}{\sigma_\varphi(f,W_{ij})}$$

$$\tau_{ij} = \langle-\rangle\mu_\varphi(f,W_{ij})$$

Thus the function that transforms the pixels belonging to the fuzzy window $W_{ij}$ takes the following form:

$$\psi_{ij}(f) = \frac{\sigma(u)}{\sigma_\varphi(f,W_{ij})}\langle\times\rangle\Big(f\langle-\rangle\mu_\varphi(f,W_{ij})\Big) \qquad (6.1.1)$$

To obtain the enhanced image $f_{enh}$, the transform  is built as a sum of the affine transforms $\psi_{ij}$ from (6.1.1), weighted with the degrees of membership $w_{ij}$:

$$f_{enh} = \sum_{j=0}^{n}\sum_{i=0}^{m} w_{ij}\langle\times\rangle\psi_{ij}(f)$$

### 6.2  The Enhancement Method for Color Images

Let be a color image described by its three scalar functions $R:\Omega\to V$, $G:\Omega\to V$, $B:\Omega\to V$. Let there be $R_{enh}$, $G_{enh}$, $B_{enh}$ the scalar components of the enhanced image. Let be $l,r,g,b$ its four components defined by relations (3.2.1 - 3.2.4). Using the formulae (5.2), (5.3) and (5.4) the fuzzy mean $\mu_\varphi(l,W_{ij})$, the fuzzy variance $\sigma_\varphi^2(l,W_{ij})$ and the fuzzy saturation $\gamma_\varphi^2(r,g,b,W_{ij})$ for the window $W_{ij}$ are computed. The fuzzy window $W_{ij}$ will supply a triple of parameters $(\lambda,\tau,\omega)$, which reflects the statistics according to the pixels belonging (in fuzzy meaning) to this window. Thus:

$$\lambda_{ij}=\frac{\sigma(u)}{\sigma_\varphi(l,W_{ij})}$$

$$\tau_{ij}=\langle-\rangle\mu_\varphi(l,W_{ij})$$

$$\omega_{ij}=\frac{\sigma(u)}{\gamma_\varphi(r,g,b,W_{ij})}$$

Thus the affine transform for window $W_{ij}$ will be defined as following:

$$R_{ij}=\lambda_{ij}\langle\times\rangle\big(l\langle+\rangle\tau_{ij}\big)\langle+\rangle\omega_{ij}\langle\times\rangle r \qquad (6.2.1)$$

$$G_{ij}=\lambda_{ij}\langle\times\rangle\big(l\langle+\rangle\tau_{ij}\big)\langle+\rangle\omega_{ij}\langle\times\rangle g \qquad (6.2.2)$$

$$B_{ij}=\lambda_{ij}\langle\times\rangle\big(l\langle+\rangle\tau_{ij}\big)\langle+\rangle\omega_{ij}\langle\times\rangle b \qquad (6.2.3)$$

The enhanced image components will be calculated using (6.2.1-6.2.3) with the following functions:

$$R_{enh}=\sum_{j=0}^{n}\sum_{i=0}^{m}w_{ij}\langle\times\rangle R_{ij}$$

$$G_{enh}=\sum_{j=0}^{n}\sum_{i=0}^{m}w_{ij}\langle\times\rangle B_{ij}$$

$$B_{enh}=\sum_{j=0}^{n}\sum_{i=0}^{m}w_{ij}\langle\times\rangle G_{ij}$$

## 7 Experimental Results

In order to exemplify the enhanced method presented in this paper, two images have been chosen.
For the first image "girl" (Fig. 1a) there was obtained the enhanced image (Fig. 1c), using a $(3\times3)$ fuzzy partition.The image was additionally processed by defining a $(3\times3)$ classical partition (Fig. 1b) and using, as well, the histogram equalization method (Fig. 1d). As for the classical partition one can notice the discontinuities between the partition windows (Fig. 1b).

The second image "aerial" is shown in Fig. 2a and its enhanced images in Fig. 2b and 2c.

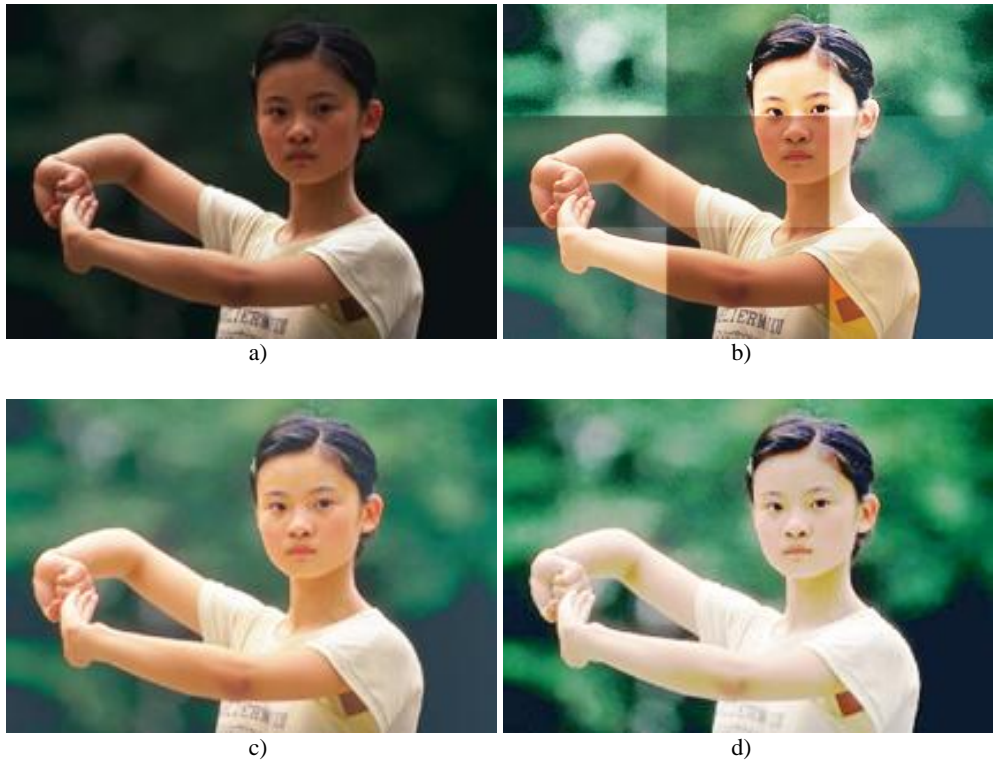

Fig. 1 - a) The original image "girl"; b) The enhanced image with (3×3) classical partition;
c) The enhanced image with (3×3) fuzzy partition;
d) The enhanced image with histogram equalization method.

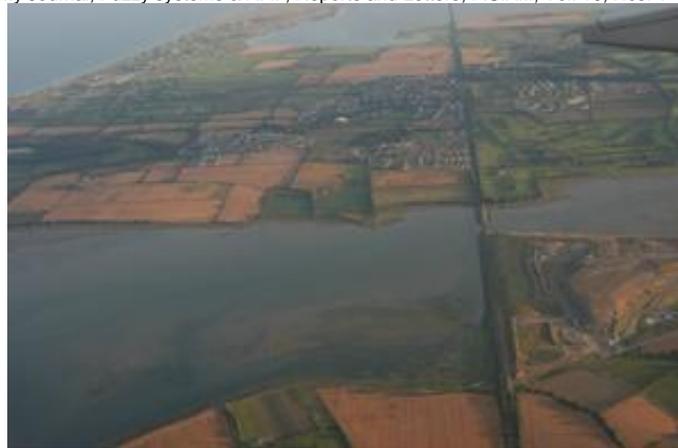

a)

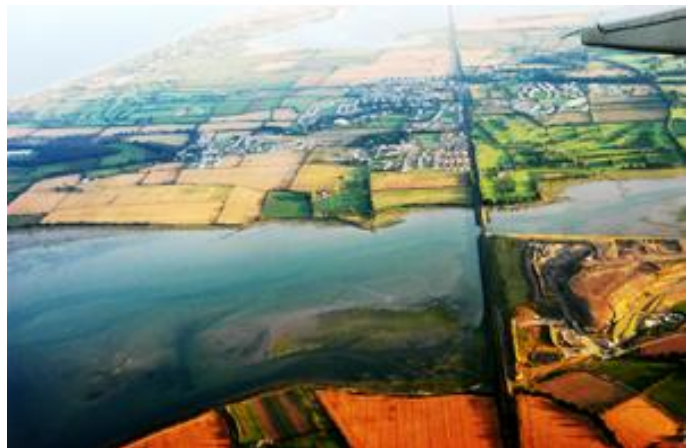

b)

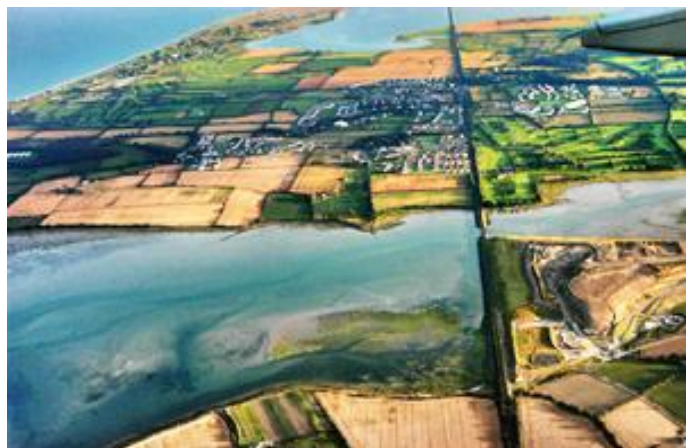

c)

Fig. 2 – a) The original image "aerial"; b) The enhanced image with histogram equalization method;
c) The enhanced image with  (10×10)  fuzzy partitions.

## 8  Conclusions

This paper presents an enhancement method for color images. This method uses simultaneously elements from the fuzzy set theory with elements from the logarithmic models theory. The calculus is made using logarithmic operations while the image structuring is made using fuzzy partitions. After splitting the image support in fuzzy windows, for each one of them an affine transform will be determined. The affine transform is the same for all color components and it is defined using three parameters that depend on brightness, contrast and saturation. In order to compute these parameters it was used a perceptual coordinate system called *lrgb*. Thus, there result different transforms depending on the brightness, the contrast and the saturation of every fuzzy window. The brightness, the contrast and the saturation dimensions can be found in the fuzzy mean, fuzzy variance and fuzzy saturation, which were computed for every element of the partition, (i.e. for every window). Using logarithmic operations defined on bounded sets permitted the avoidance of the truncation operations. The fuzzy partition of the image support allows building affine transforms adjusted to diverse areas of the processed image. As we can observe in the experimental result section, the enhanced images obtained by the new method are better than those obtained by histogram equalization method.

## References

[1] V. PĂTRAŞCU, V. BUZULOIU: *Color Image Enhancement in the Framework of Logarithmic Models*, The 8[th] IEEE International Conference on Telecommunications, Vol. **1**, ICT'01, Bucharest, Romania, pp. 199-204, 2001.

[2] V. PĂTRAŞCU, V. BUZULOIU: *A Mathematical Model for Logarithmic Image Processing*, The 5[th] World Multi-Conference on Systemics, Cybernetics and Informatics, Vol **13**, SCI'01, Orlando, USA, pp. 117-122, 2001.

[3] V. PĂTRAŞCU, V. BUZULOIU: *Modelling of Histogram Equalisation with Logarithmic Affine Transforms*, Recent Trends in Multimedia Information Processing, (Ed. P. Liatsis), World Scientific Press, 2002, Proceedings of the 9[th] International Workshop on Systems, Signals and Image Processing, IWSSIP'02, Manchester, United Kingdom, pp 312-316, 2002.

[4] V. PĂTRAŞCU: *Gray Level Image Enhancement using the Support Fuzzification in the Framework of Logarithmic Models*, In proceedings of the International Conference Model-based Imaging, Rendering, Image Analysis and Graphical Special Effects, Mirage'03, INRIA Rocquencourt, France, pp. 123-128, 2003.

[5] V. PĂTRAŞCU: *Color Image Enhancement Using the Support Fuzzification*, In Fuzzy Sets and Systems – IFSA'03, Vol. **LNAI 2715**, (Eds. T. Bilgiç, B. De Baets, O. Kaynak), Springer-Verlag Berlin Heidelberg 2003, Proceedings of the 10[th] International Fuzzy Systems Association World Congress, Istanbul, Turkey, pp. 412-419, 2003.

[6] V. PĂTRAŞCU: *Gray level image enhancement method using the logarithmic model*, Acta Tehnica Napocensis, Electronics and Telecommunications, Vol. **44**, Nr. 2, 2003, Cluj-Napoca, Romania, pp. 39-50, 2003.

[7] V. PĂTRAŞCU: *Gray level image processing using algebraic structures*, Proceedings of the 11[th] Conference on applied and industrial mathematics – CAIM'03, Oradea, Romania, pp. 167-172, 2003.

[8] V. PĂTRAŞCU: *Color Image Enhancement Using the Separated Modification of Luminosity and Saturation*, Proceedings of the Conference "Modern technologies in the XXI century", Military Technical Academy, Bucharest, Romania, 2003.